\documentclass[11pt,a4paper]{article}
\usepackage[hyperref]{acl2018}
\usepackage{times}
\usepackage{latexsym}
\usepackage{CJK}
\usepackage{amsmath}
\usepackage[ruled]{algorithm2e}
\usepackage{url}
\usepackage{graphicx}
\usepackage{multirow}
\usepackage{color}
\usepackage{subfigure}
\usepackage{amssymb}
\usepackage{enumitem}

\usepackage{footnote}
\makesavenoteenv{figure}

\aclfinalcopy 

\newcommand*{\Scale}[2][4]{\scalebox{#1}{$#2$}}%

\title{Chinese NER Using Lattice LSTM}

\newcommand*\samethanks[1][\value{footnote}]{\footnotemark[#1]}
\author{Yue Zhang\thanks{\quad Equal contribution.} \and Jie Yang\samethanks\\ 
  Singapore University of Technology and Design \\
  {\tt yue\_zhang@sutd.edu.sg} \\
  {\tt jie\_yang@mymail.sutd.edu.sg} \\}

\date{}

\begin{document}
\begin{CJK*}{UTF8}{gbsn}
\maketitle
\begin{abstract}
We investigate a lattice-structured LSTM model for Chinese NER, which encodes a sequence of input characters as well as all potential words that match a lexicon. Compared with character-based methods, our model explicitly leverages word and word sequence information. Compared with word-based methods, lattice LSTM does not suffer from segmentation errors. Gated recurrent cells allow our model to choose the most relevant characters and words from a sentence for better NER results. Experiments on various datasets show that lattice LSTM outperforms both word-based and character-based LSTM baselines, achieving the best results. 
\end{abstract}

\section{Introduction}

As a fundamental task in information extraction, named entity recognition (NER) has received constant research attention over the recent years. The task has traditionally been solved as a sequence labeling problem, where entity boundary and category labels are jointly predicted. The current state-of-the-art for English NER has been achieved by using LSTM-CRF models \cite{lample2016neural,ma2016end,chiu2015named,liu2017empower} with character information being integrated into word representations.

Chinese NER is correlated with word segmentation. In particular, named entity boundaries are also word boundaries. One intuitive way of performing Chinese NER is to perform word segmentation first, before applying word sequence labeling. The segmentation $\rightarrow$ NER pipeline, however, can suffer the potential issue of error propagation, since NEs are an important source of OOV in segmentation, and incorrectly segmented entity boundaries lead to NER errors. This problem can be severe in the open domain since cross-domain word segmentation remains an unsolved problem \cite{liu2012unsupervised,jiang2013discriminative,liu2014domain,qiu2015word,chen2017adversarial,huang2017addressing}. 
It has been shown that character-based methods outperform word-based methods for Chinese NER \cite{he2008chinese,liu2010chinese,li2014comparison}. 

\begin{figure}[!tp] 
  \centering 
  \includegraphics[width=3.0in]{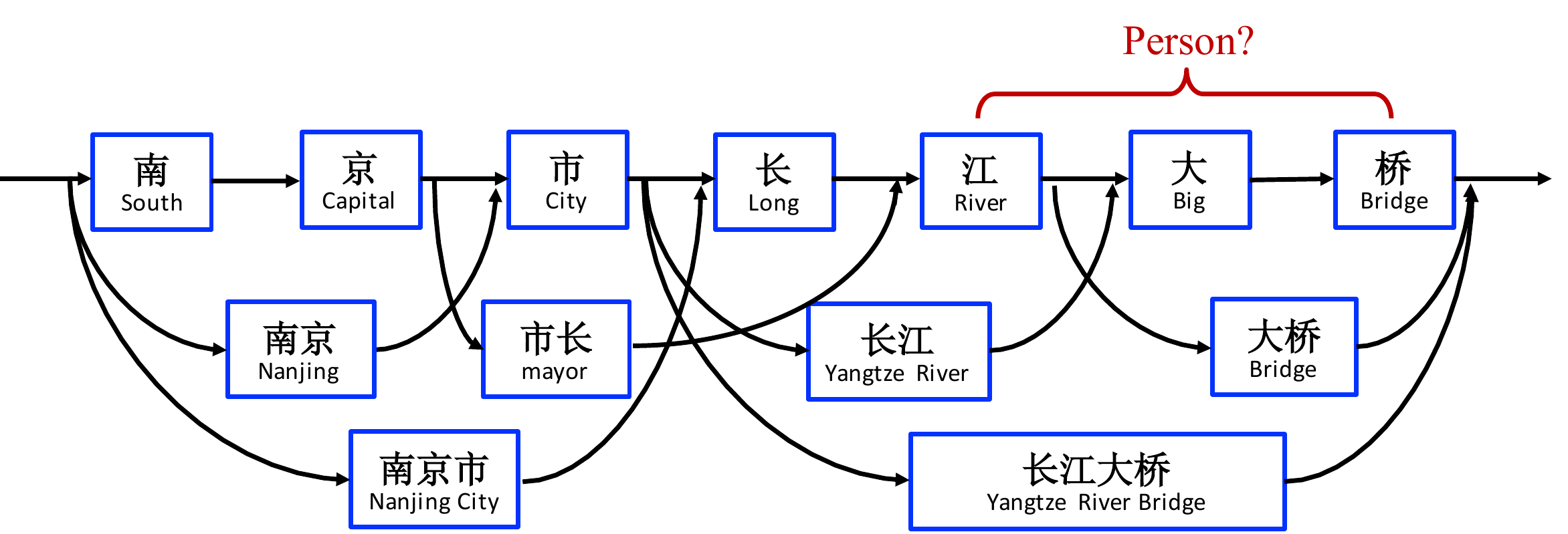}
  \vspace*{-5mm}
  \caption{Word character lattice.}
  \label{fig:latticebasic}
\end{figure}
One drawback of character-based NER, however, is that explicit word and word sequence information is not fully exploited, which can be potentially useful. To address this issue, we integrate latent word information into character-based LSTM-CRF by representing lexicon words from the sentence using a lattice structure LSTM. As shown in Figure \ref{fig:latticebasic}, we construct a word-character lattice by matching a sentence with a large automatically-obtained lexicon. As a result, word sequences such as ``长江大桥\,(Yangtze River Bridge)'', ``长江\,(Yangtze River)'' and ``大桥\,(Bridge)'' can be used to disambiguate potential relevant named entities in a context, such as the person name ``江大桥\,(Daqiao Jiang)''. 

\begin{figure}[!tp] 
  \centering 
  \includegraphics[width=3.0in]{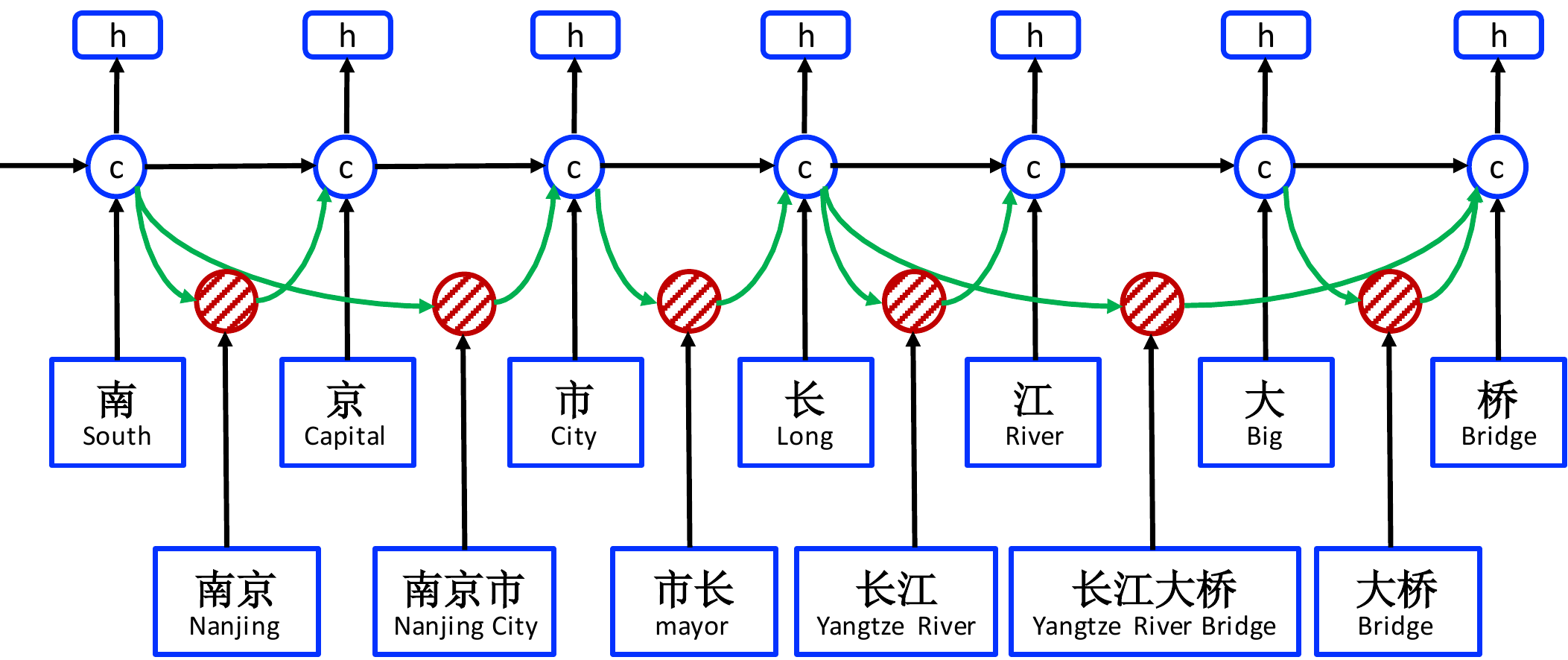}
  \vspace*{-5mm}
  \caption{Lattice LSTM structure.}
  \label{fig:latticestructure}
\end{figure}

Since there are an exponential number of word-character paths in a lattice, we leverage a lattice LSTM structure for automatically controlling information flow from the beginning of the sentence to the end.  As shown in Figure \ref{fig:latticestructure}, gated cells are used to dynamically route information from different paths to each character. Trained over NER data, the lattice LSTM can learn to find more useful words from context automatically for better NER performance. Compared with character-based and word-based NER methods, our model has the advantage of leveraging explicit word information over character sequence labeling without suffering from segmentation error. 

Results show that our model significantly outperforms both character sequence labeling models and word sequence labeling models using LSTM-CRF, giving the best results over a variety of Chinese NER datasets across different domains. Our code and data are released at \url{https://github.com/jiesutd/LatticeLSTM}.

\section{Related Work}
Our work is in line with existing methods using neural network for NER. \citet{hammerton2003named} attempted to solve the problem using a uni-directional LSTM, which was among the first neural models for NER. \citet{collobert2011natural} used a CNN-CRF structure, obtaining competitive results to the best statistical models. \citet{dos2015boosting} used character CNN to augment a CNN-CRF model. Most recent work leverages an LSTM-CRF architecture. \citet{huang2015bidirectional} uses hand-crafted spelling features; \citet{ma2016end} and \citet{chiu2015named} use a character CNN to represent spelling characteristics; \citet{lample2016neural} use a character LSTM instead. Our baseline word-based system takes a similar structure to this line of work. 

Character sequence labeling has been the dominant approach for Chinese NER \cite{chen2006chinesener,lu2016multi,dong2016character}. There have been explicit discussions comparing statistical word-based and character-based methods for the task, showing that the latter is empirically a superior choice \cite{he2008chinese,liu2010chinese,li2014comparison}. We find that with proper representation settings, the same conclusion holds for neural NER. On the other hand, lattice LSTM is a better choice compared with both word LSTM and character LSTM.

How to better leverage word information for Chinese NER has received continued research attention \cite{gao2005chinese}, where segmentation information has been used as soft features for NER \cite{zhao2008unsupervised,peng2015named,he2017f}, and joint segmentation and NER has been investigated using dual decomposition \cite{xu2014joint}, multi-task learning \cite{peng2016improving}, etc. Our work is in line, focusing on neural representation learning. While the above methods can be affected by segmented training data and segmentation errors, our method does not require a word segmentor. The model is conceptually simpler by not considering multi-task settings. 

External sources of information has been leveraged for NER. In particular, lexicon features have been widely used \cite{collobert2011natural,passos2014lexicon,huang2015bidirectional,luo2015joint}. \citet{rei2017semi} uses a word-level language modeling objective to augment NER training, performing multi-task learning over large raw text. \citet{peters2017semi} pretrain a character language model to enhance word representations. \citet{yang2017transfer} exploit cross-domain and cross-lingual knowledge via multi-task learning.  We leverage external data by pretraining word embedding lexicon over large automatically-segmented texts, while semi-supervised techniques such as language modeling are orthogonal to and can also be used for our lattice LSTM model.

Lattice structured RNNs can be viewed as a natural extension of tree-structured RNNs \cite{tai-socher-manning:2015:ACL-IJCNLP} to DAGs. They have been used to model motion dynamics \cite{Sun_2017_ICCV}, dependency-discourse DAGs \cite{DBLP:journals/tacl/PengPQTY17}, as well as speech tokenization lattice \cite{sperber-EtAl:2017:EMNLP2017} and multi-granularity segmentation outputs \cite{DBLP:conf/aaai/SuTXJSL17} for NMT encoders. Compared with existing work, our lattice LSTM is different in both motivation and structure. For example, being designed for character-centric lattice-LSTM-CRF sequence labeling, it has recurrent cells but not hidden vectors for words. To our knowledge, we are the first to design a novel lattice LSTM representation for mixed characters and lexicon words, and the first to use a word-character lattice for segmentation-free Chinese NER.

\section{Model}

We follow the best English NER model \cite{huang2015bidirectional,ma2016end,lample2016neural}, using LSTM-CRF as the main network structure. Formally, denote an input sentence as $s=c_1,c_2,\ldots,c_m$, where $c_j$ denotes the $j$th character. $s$ can further be seen as a word sequence $s=w_1,w_2,\ldots,w_n$, where $w_i$ denotes the $i$th word in the sentence, obtained using a Chinese segmentor.  We use $t(i, k)$ to denote the index $j$ for the $k$th character in the $i$th word in the sentence. Take the sentence in Figure \ref{fig:latticebasic} for example. If the segmentation is ``南京市\;\;长江大桥'', and indices are from 1, then $t(2, 1)=4$ (长) and $t(1, 3)=3$ (市). We use the BIOES tagging scheme \cite{ratinov2009design} for both word-based and character-based NER tagging.

\subsection{Character-Based Model}\label{sec:charmodel}
\begin{figure}[!t]
  \centering 
  \subfigure[Character-based model.]{ 
    \label{fig:charbase} 
    \includegraphics[width=3in]{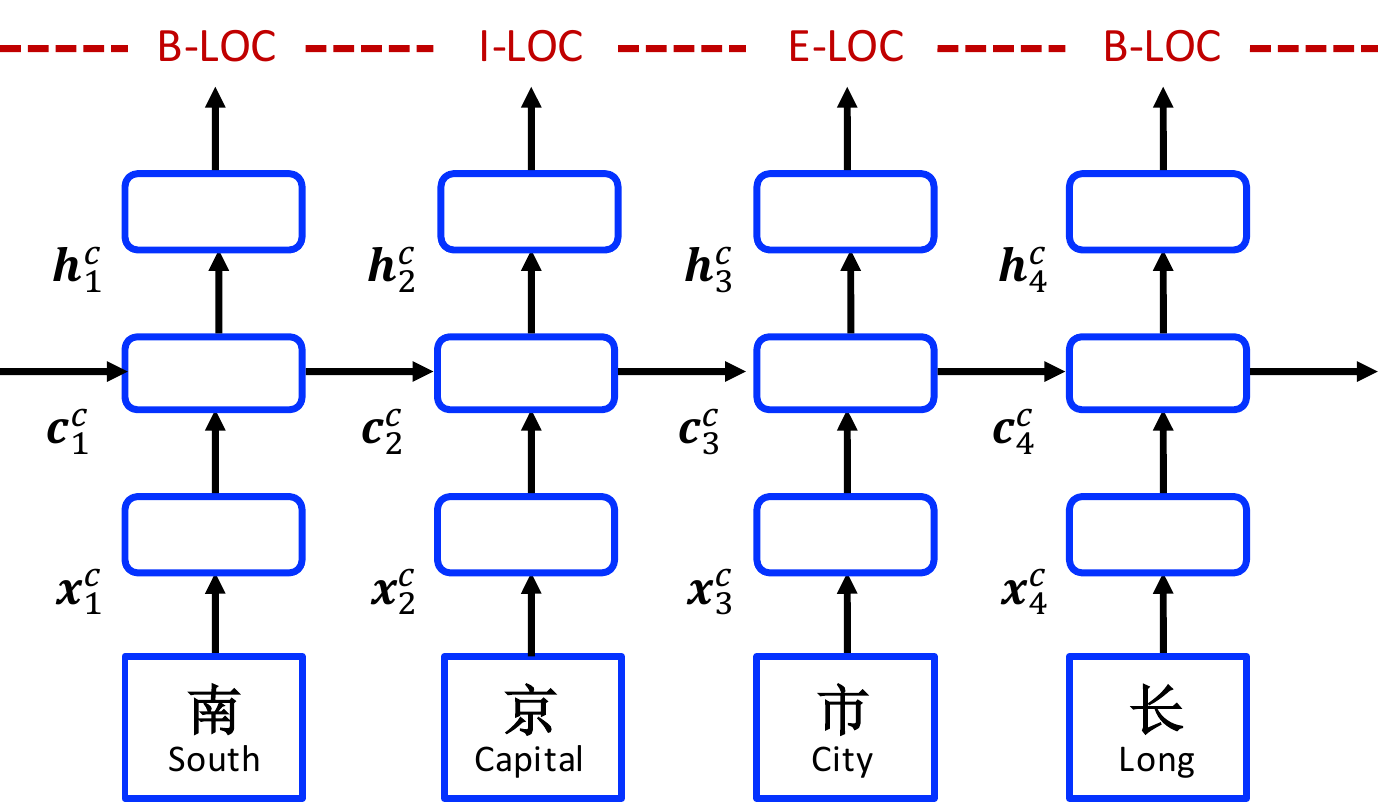}} 
  \subfigure[Word-based model.]{ 
    \label{fig:wordbase} 
    \includegraphics[width=3in]{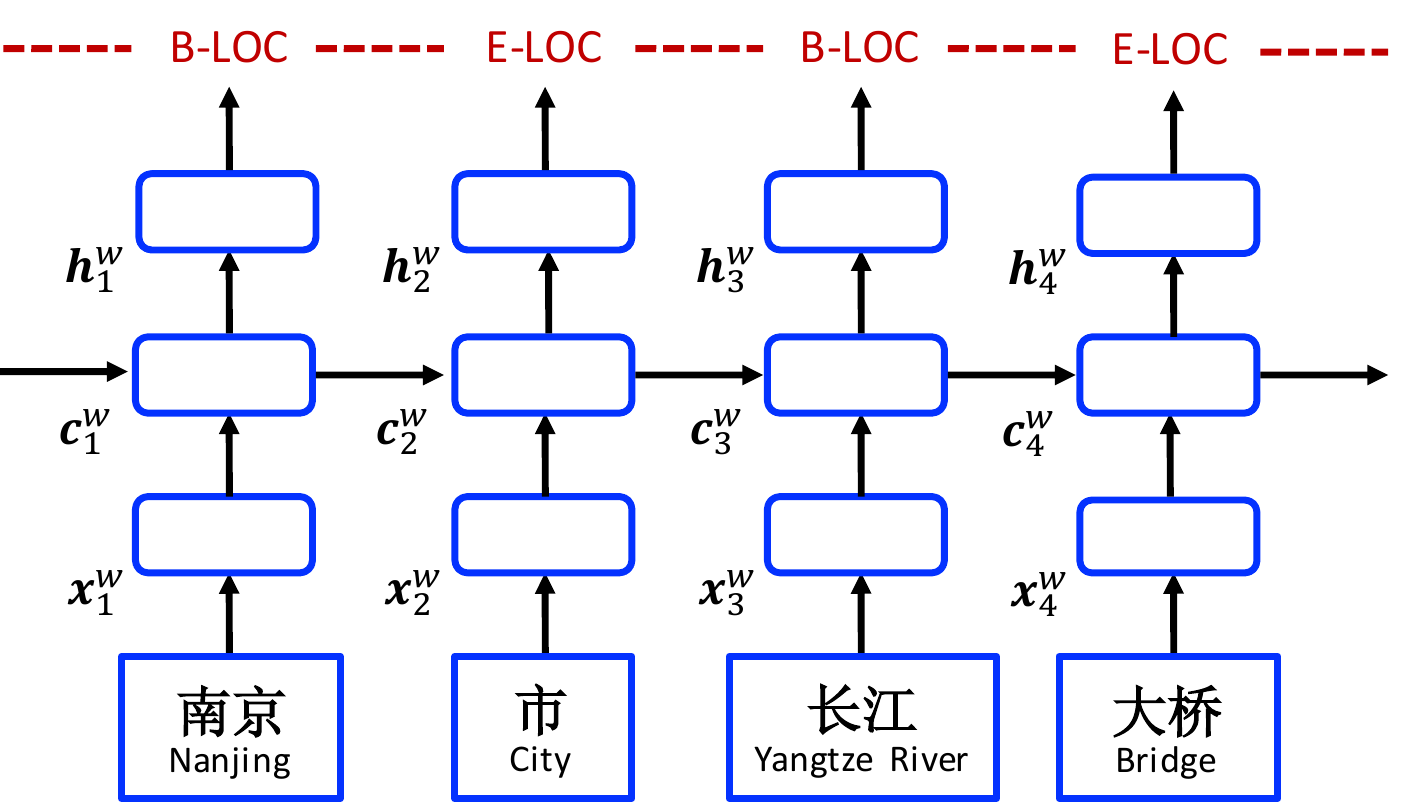}} 
  \subfigure[Lattice model.]{ 
    \label{fig:latticebase} 
    \includegraphics[width=3in]{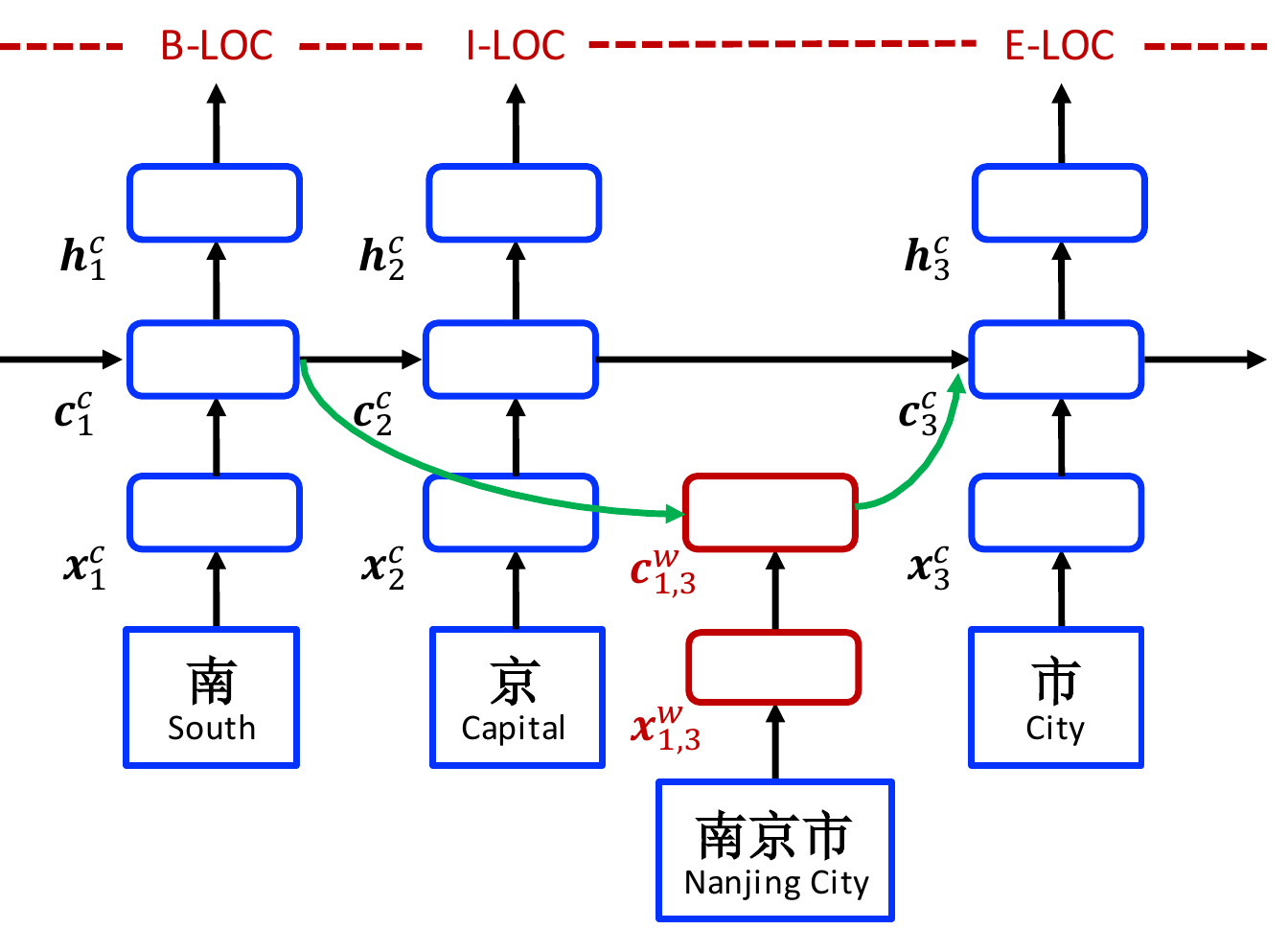}} 
  \caption{Models.\footnote{To keep the figure concise, we (i) do not show gate cells, which uses $\textbf{h}_{t-1}$ for calculating $\textbf{c}_t$; (ii) only show one direction.}} 
  \label{fig:models} 
\end{figure}

The character-based model is shown in Figure \ref{fig:charbase}. It uses an LSTM-CRF model on the character sequence $c_1,c_2,\ldots,c_m$. Each character $c_j$ is represented using
\begin{equation}
\textbf{x}^c_j = \textbf{e}^c(c_j)
\end{equation}
$\textbf{e}^c$ denotes a character embedding lookup table. 

A bidirectional LSTM (same structurally as Eq. \ref{eq:clstm}) is applied to $\textbf{x}_1, \textbf{x}_2, \ldots, \textbf{x}_m$ to obtain $\overrightarrow{\textbf{h}}^c_1, \overrightarrow{\textbf{h}}^c_2, \ldots, \overrightarrow{\textbf{h}}^c_m$ and $\overleftarrow{\textbf{h}}^c_1, \overleftarrow{\textbf{h}}^c_2, \ldots, \overleftarrow{\textbf{h}}^c_m$ in the left-to-right and right-to-left directions, respectively, with two distinct sets of parameters. The hidden vector representation of each character is:
\begin{equation}
\textbf{h}^c_j = [\overrightarrow{\textbf{h}}^c_j;\overleftarrow{\textbf{h}}^c_j]
\end{equation}

A standard CRF model (Eq. \ref{eq:crf}) is used on $\textbf{h}^c_1, \textbf{h}^c_2,\ldots,\textbf{h}^c_m$ for sequence labelling.

\noindent \textbullet \; \textbf{Char + bichar}. Character bigrams have been shown useful for representing characters in word segmentation \cite{chen-EtAl:2015:EMNLP2,yang-zhang-dong:2017:Long}. We augment the character-based model with bigram information by concatenating bigram embeddings with character embeddings:
\begin{equation}
\textbf{x}^c_j=[\textbf{e}^c(c_j); \textbf{e}^b(c_j, c_{j+1})],
\end{equation}
where $\textbf{e}^b$ denotes a charater bigram lookup table.

\noindent \textbullet \; \textbf{Char + softword}. It has been shown that using segmentation as soft features for character-based NER models can lead to improved performance \cite{zhao2008unsupervised,peng2016improving}. We augment the character representation with segmentation information by concatenating segmentation label embeddings to character embeddings:
\begin{equation}
\textbf{x}^c_j = [\textbf{e}^c(c_j); \textbf{e}^s(\textit{seg}(c_j))],
\end{equation}

where $\textbf{e}^s$ represents a segmentation label embedding lookup table. \textit{seg}$(c_j)$ denotes the segmentation label on the character $c_j$ given by a word segmentor. We use the BMES scheme for representing segmentation \cite{xue2003Chinese}.

\begin{equation}
\textbf{h}_i^w = [\overrightarrow{\textbf{h}_i^w};\overleftarrow{\textbf{h}_i^w}]
\end{equation}
Similar to the character-based case, a standard CRF model (Eq. \ref{eq:crf}) is used on $\textbf{h}^w_1, \textbf{h}^w_2,\ldots,\textbf{h}^w_m$ for sequence labelling.

\subsection{Word-Based Model}\label{sec:word-based}
The word-based model is shown in Figure \ref{fig:wordbase}. It takes the word embedding $\textbf{e}^w(w_i)$ for representation each word $w_i$:
\begin{equation}\label{eq:xi}
\textbf{x}^w_i = \textbf{e}^w(w_i),
\end{equation}
where $\textbf{e}^w$ denotes a word embedding lookup table.
A bi-directioanl LSTM (Eq. \ref{eq:clstm}) is used to obtain a left-to-right sequence of hidden states $\overrightarrow{\textbf{h}_1^w}, \overrightarrow{\textbf{h}_2^w}, \ldots, \overrightarrow{\textbf{h}_n^w}$ and a right-to-left sequence of hidden states $\overleftarrow{\textbf{h}_1^w}, \overleftarrow{\textbf{h}_2^w}, \ldots, \overleftarrow{\textbf{h}_n^w}$ for the words $w_1, w_2, \ldots, w_n$, respectively. Finally, for each word $w_i$, $\overrightarrow{\textbf{h}_i^w}$ and $\overleftarrow{\textbf{h}_i^w}$ are concatenated as its representation:

\textbf{Integrating character representations}

Both character CNN \cite{ma2016end} and LSTM \cite{lample2016neural} have been used for representing the character sequence within a word. We experiment with both for Chinese NER. Denoting the representation of characters within $w_i$ as $\textbf{x}^c_i$, a new word representation is obtained by concatenation of $\textbf{e}^w(w_i)$ and $\textbf{x}^c_i$:
\begin{equation}
\textbf{x}^w_i = [\textbf{e}^w(w_i); \textbf{x}^c_i]
\end{equation}

\noindent  \textbullet \; \textbf{Word + char LSTM}. Denoting the embedding of each input character as $\textbf{e}^c(c_j)$, we use a bi-directional LSTM (Eq. \ref{eq:clstm}) to learn hidden states $\overrightarrow{\textbf{h}}^c_{t(i, 1)}, \ldots, \overrightarrow{\textbf{h}}^c_{t(i, \textit{len}(i))}$ and $\overleftarrow{\textbf{h}}^c_{t(i, 1)}, \ldots, \overleftarrow{\textbf{h}}^c_{t(i, \textit{len}(i))}$ for the characters $c_{t(i, 1)}, \ldots, c_{t(i,\textit{len}(i))}$ of $w_i$, where \textit{len}$(i)$ denotes the number of characters in $w_i$. The final character representation for $w_i$ is:
\begin{equation}\label{eq:xci}
\textbf{x}^c_i=[\overrightarrow{\textbf{h}}^c_{t(i,{\it len}(i))}; \overleftarrow{\textbf{h}}^c_{t(i, 1)}]
\end{equation}

\noindent \textbullet \; \textbf{Word + char LSTM$'$}. We investigate a variation of word + char LSTM model that uses a \textbf{single} LSTM to obtain $\overrightarrow{\textbf{h}}^c_j$ and $\overleftarrow{\textbf{h}}^c_j$ for each $c_j$. It is similar with the structure of \newcite{liu2017empower} but not uses the highway layer. The same LSTM structure as defined in Eq. \ref{eq:clstm} is used, and the same method as Eq. \ref{eq:xci} is used to integrate character hidden states into word representations. 

\noindent \textbullet \; \textbf{Word + char CNN}. A standard CNN \cite{lecun1989backpropagation} structure is used on the character sequence of each word to obtain its character representation $\textbf{x}^c_i$. Denoting the embedding of character $c_j$ as $\textbf{e}^c(c_j)$, the vector $\textbf{x}^c_i$ is given by:
\begin{equation}
\Scale[0.87]{
\textbf{x}^c_i=\max \limits_{t(i,1)\le j\le t(i,len(i))}(\textbf{W}^\top_{\textsc{Cnn}}
{\left[ \begin{array}{c}
\textbf{e}^c(c_{j-\frac{ke-1}{2}}) \\
\ldots \\
\textbf{e}^c(c_{j+\frac{ke-1}{2}})\\ 
\end{array} 
\right ]} 
 + \textbf{b}_{\textsc{Cnn}})
 },
\end{equation}
where $\textbf{W}_{\textsc{Cnn}}$ and $\textbf{b}_{\textsc{Cnn}}$ are parameters, $ke=3$ is the kernal size and $max$ denotes max pooling.

\subsection{Lattice Model}
The overall structure of the word-character lattice model is shown in Figure \ref{fig:latticestructure}, which can be viewed as an extension of the character-based model, integrating word-based cells and additional gates for controlling information flow.  

Shown in Figure \ref{fig:latticebase}, the input to the model is a character sequence $c_1,c_2,\ldots,c_m$, together with all character subsequences that match words in a lexicon $\mathbb{D}$. As indicated in Section 2, we use automatically segmented large raw text for buinding $\mathbb{D}$. Using $w^d_{b,e}$ to denote such a subsequence that begins with character index $b$ and ends with character index $e$, the segment $w^d_{1, 2}$ in Figure \ref{fig:latticebasic} is ``南京\;(Nanjing)'' and $w^d_{7,8}$ is ``大桥\;(Bridge)''. 

Four types of vectors are involved  in the model, namely \textit{input vectors}, \textit{output hidden vectors}, \textit{cell vectors} and \textit{gate vectors}. As basic components, a character input vector is used to represent each chacracter $c_j$ as in the character-based model:
\begin{equation}
\textbf{x}^c_j = \textbf{e}^c(c_j)
\end{equation}

The basic recurrent structure of the model is constructed using a character cell vector $\textbf{c}^c_j$ and a hidden vector $\textbf{h}_j^c$ on each $c_j$, where $\textbf{c}^c_j$ serves to record recurrent information flow from the beginning of the sentence to $c_j$ and $\textbf{h}_j^c$ is used for CRF sequence labelling using Eq. \ref{eq:crf}. 

The basic recurrent LSTM functions are: \\
\begin{equation}\label{eq:clstm}
\Scale[0.87]{
\begin{aligned}
{\left[ \begin{array}{c}
\textbf{i}_j^c \\
\textbf{o}_j^c\\
\textbf{f}_j^c\\
\boldsymbol{\widetilde{c}}_j^c
\end{array} 
\right ]} 
&=
{\left[ \begin{array}{c}
\sigma \\
\sigma\\
\sigma\\
tanh 
\end{array} 
\right ]} 
\Big(
\textbf{W}^{c\top}
{\left[ \begin{array}{c}
\textbf{x}_j^c \\
\textbf{h}_{j-1}^c\\ 
\end{array} 
\right ]} 
 +
\textbf{b}^c
 \Big)
\\
\textbf{c}_j^c &= \textbf{f}_j^c\odot \textbf{c}_{j-1}^c + \textbf{i}_j^c \odot \boldsymbol{\widetilde{c}}_j^c \\
\textbf{h}_j^c &= \textbf{o}_j^c\odot tanh(\textbf{c}_j^c) \\
\end{aligned}
}
\end{equation}
where $\textbf{i}_j^c, \textbf{f}_j^c$ and $\textbf{o}_j^c$ denote a set of input, forget and output gates, respectively. $\textbf{W}^{c\top}$ and $\textbf{b}^c$ are model parameters. $\sigma()$ represents the sigmoid function.

Different from the character-based model, however, the computation of $\textbf{c}^c_j$ now considers lexicon subsequences $w^d_{b, e}$ in the sentence. In particular, each subsequence $w^d_{b, e}$ is represented using
\begin{equation}
\textbf{x}^w_{b, e} = \textbf{e}^w(w^d_{b, e}),
\end{equation}
where $\textbf{e}^w$ denotes the same word embedding lookup table as in Section~\ref{sec:word-based}. 

In addition, a word cell $\textbf{c}^w_{b, e}$ is used to represent the recurrent state of $\textbf{x}^w_{b, e}$ from the beginning  of the sentence. The value of $\textbf{c}^w_{b, e}$ is calculated by:

\begin{equation}
\Scale[0.9]{
\begin{aligned}
{\left[ \begin{array}{c}
\textbf{i}_{b,e}^w \\
\textbf{f}_{b,e}^w\\
\boldsymbol{\widetilde{c}}_{b,e}^w 
\end{array} 
\right ]} 
&=
{\left[ \begin{array}{c}
\sigma \\
\sigma\\
tanh 
\end{array} 
\right ]} 
\Big(
\textbf{W}^{w\top}
{\left[ \begin{array}{c}
\textbf{x}^w_{b, e} \\
\textbf{h}_b^c\\ 
\end{array} 
\right ]} 
 +
\textbf{b}^w
 \Big)
\label{eq:word}\\
\textbf{c}^w_{b,e} &= \textbf{f}_{b, e}^w\odot \textbf{c}^c_b + \textbf{i}_{b, e}^w \odot \boldsymbol{\widetilde{c}}_{b, e}^w \\
\end{aligned}
}
\end{equation}
where $\textbf{i}_{b, e}^w$ and $\textbf{f}_{b, e}^w$ are a set of input and forget gates. There is no output gate for word cells since labeling is performed only at the character level.

With $\textbf{c}^w_{b,e}$, there are more recurrent paths for information flow into each $\textbf{c}^c_j$. For example, in Figure 2, input sources for $\textbf{c}^c_7$ include $\textbf{x}^c_7$ (桥\;Bridge), $\textbf{c}^w_{6,7}$ (大桥\;Bridge) and $\textbf{c}^w_{4,7}$ (长江大桥\;Yangtze River Bridge).\footnote{We experimented with alternative configurations on indexing word and character path links, finding that this configuration gives the best results in preliminary experiments. Single-character words are excluded; the final performance drops slightly after integrating single-character words.} We link all $\textbf{c}_{b,e}^w$ with $b\in \{b'|w^d_{b',e}\in \mathbb{D}\}$ to the cell $\textbf{c}_e^c$.
We use an additional gate $\textbf{i}^c_{b,e}$ for each subsequence cell $\textbf{c}^w_{b,e}$ for controlling its contribution into $\textbf{c}^c_{b,e}$:
\begin{equation}
\Scale[0.9]{
 \label{eq:gw}\\
\textbf{i}_{b,e}^c = \sigma
\big(
\textbf{W}^{l\top}{\left[ \begin{array}{c}
\textbf{x}_{e}^c\\
\textbf{c}_{b,e}^w\\ 
\end{array} 
\right ]} 
+\textbf{b}^l
\big) \\
}
\end{equation}

The calculation of cell values $\textbf{c}^c_j$ thus becomes
\begin{equation}\label{eq:avgsum}
\Scale[0.9]{
\textbf{c}_j^c =  \sum\limits_{b\in \{b'|w^d_{b',j} \in \mathbb{D}\}} \boldsymbol\alpha_{b,j}^{c}\odot \boldsymbol{c}_{b,j}^{w} + \boldsymbol\alpha^c_j\odot \boldsymbol{\widetilde{c}}_j^c 
}
\end{equation}

In Eq. \ref{eq:avgsum}, the gate values $\textbf{i}_{b,j}^c$ and $\textbf{i}^c_j$ are normalised to $\boldsymbol\alpha_{b,j}^{c}$ and $\boldsymbol\alpha^{c}_j$ by setting the sum to $\textbf{1}$.
\begin{equation} \label{eq:norm}
\Scale[0.88]{
\begin{aligned}
\boldsymbol\alpha_{b,j}^{c} &= \dfrac{exp(\textbf{i}_{b,j}^c)}{exp(\textbf{i}^c_j)+\sum_{b'\in \{b''|w^d_{b'',j} \in \mathbb{D}\}}exp(\textbf{i}_{b',j}^c)} \\
\boldsymbol\alpha^{c}_j &= \dfrac{exp(\textbf{i}^c_j)}{exp(\textbf{i}^c_j)+\sum_{b'\in \{b''|w^d_{b'',j} \in \mathbb{D}\}}exp(\textbf{i}_{b',j}^c)} \\
\end{aligned}
}
\end{equation}

The final hidden vectors $\textbf{h}_j^c$ are still computed as described by Eq. \ref{eq:clstm}. During NER training, loss values back-propagate to the parameters $\textbf{W}^c, \textbf{b}^c, \textbf{W}^w, \textbf{b}^w, \textbf{W}^l$ and $\textbf{b}^l$ allowing the model to dynamically focus on more relevant words during NER labelling.

\subsection{Decoding and Training}
A standard CRF layer is used on top of $\textbf{h}_1,\textbf{h}_2,\ldots,\textbf{h}_{\tau}$, where $\tau$ is $n$ for character-based and lattice-based models and $m$ for word-based models. The probability of a label sequence $y=l_1,l_2,\ldots,l_{\tau}$ is
\begin{equation}\label{eq:crf}
\Scale[0.9]{
P(y|s)=\dfrac{exp(\sum_{i}(\textbf{W}_\textsc{Crf}^{l_i}\textbf{h}_i + b_\textsc{Crf}^{(l_{i-1},l_i)}))}{\sum_{y^\prime}exp(\sum_{i}(\textbf{W}_\textsc{Crf}^{l_i^\prime}\textbf{h}_i + b_\textsc{Crf}^{(l_{i-1}^\prime, l_i^\prime)}))}
}
\end{equation}
Here $y^\prime$ represents an arbitary label sequence, and $\textbf{W}_\textsc{Crf}^{l_i}$ is  a model parameter specific to $l_i$, and $b_\textsc{Crf}^{(l_{i-1},l_i)}$ is a bias specific to $l_{i-1}$ and $l_i$.

We use the first-order Viterbi algorithm to find the highest scored label sequence over a word-based or character-based input sequence. Given a set of manually labeled training data $\{(s_i, y_i)\}|^N_{i=1}$, sentence-level log-likelihood loss with $L_2$ regularization is used to train the model:
\begin{equation}
\Scale[0.9]{
 L = \sum_{i=1}^{N}log(P(y_i|s_i)) + \frac{\lambda}{2} ||\Theta||^2,
 }
\end{equation}
where $\lambda$ is the $L_2$ regularization parameter and $\Theta$ represents the parameter set.

\begin{table}[!tp]
\begin{center}
\scalebox{0.7}{
\begin{tabular}{|l|l|l|l|l|}
\hline 
\textbf{Dataset} &\textbf{Type}&\textbf{Train} & \textbf{Dev} & \textbf{Test} \\ 
\hline
\multirow{2}*{OntoNotes} 
&Sentence& 15.7k & 4.3k &4.3k \\
&Char & 491.9k  &200.5k &208.1k\\
\hline
\multirow{2}*{MSRA} 
&Sentence& 46.4k & -- &4.4k \\
&Char & 2169.9k  &-- &172.6\\
\hline
\multirow{2}*{Weibo} 
&Sentence& 1.4k & 0.27k &0.27k \\
&Char & 73.8k  &14.5k &14.8k \\
\hline
\multirow{2}*{resume} 
&Sentence& 3.8k & 0.46k &0.48k \\
&Char & 124.1k  &13.9k & 15.1k \\
\hline
\end{tabular}
}
\end{center}
\caption{Statistics of datasets.}
\label{tab:OverallSta}
\end{table}

\section{Experiments} 
We carry out an extensive set of experiments to investigate the effectiveness of word-character lattice LSTMs across different domains. In addition, we aim to empirically compare word-based and character-based neural Chinese NER under different settings. Standard precision (P), recall (R) and F1-score (F1) are used as evaluation metrics.

\subsection{Experimental Settings}
\textbf{Data}. Four datasets are used in this paper, which include OntoNotes 4 \cite{weischedel2011ontonotes}, MSRA \cite{levow2006third} Weibo NER \cite{peng2015named,he2017f} and a Chinese resume dataset that we annotate. Statistics of the datasets are shown in Table \ref{tab:OverallSta}. We take the same data split as \citet{che2013named} on OntoNotes. The development set of OntoNotes is used for reporting development experiments. While the OntoNotes and MSRA datasets are in the news domain, the Weibo NER dataset is drawn from the social media website Sina Weibo.\footnote{\url{https://www.weibo.com/}}

\begin{table}[!tp]
\begin{center}
\scalebox{0.9}{
\begin{tabular}{|l|l|l|l|}
\hline \textbf{Statistics} &  \textbf{Train} & \textbf{Dev} & \textbf{Test}\\ \hline
Country&260  &33 & 28\\
Educational Institution&858  &106 & 112\\
Location&47  &2 & 6\\
Personal Name&952  &110 & 112\\
Organization&4611  &523 & 553\\
Profession&287  &18 & 33\\
Ethnicity Background&115  &15 & 14\\
Job Title&6308  &690 & 772\\
\hline
Total Entity & 13438  &1497 & 1630\\
\hline
\end{tabular}
}
\end{center}
\caption{Detailed statistics of resume NER.}
\label{tab:ResumeSta}
\end{table}

For more variety in test domains, we collected a resume dataset from Sina Finance\footnote{\url{http://finance.sina.com.cn/stock/index.shtml}}, which consists of resumes of senior executives from listed companies in the Chinese stock market. We randomly selected 1027 resume summaries and manually annotated 8 types of named entities with YEDDA system \cite{yang2017yedda}. Statistics of the dataset is shown in Table \ref{tab:ResumeSta}. The inter-annotator agreement is 97.1\%. We release this dataset as a resource for further research.

\textbf{Segmentation}. For the OntoNotes and MSRA datasets, gold-standard segmentation is available in the training sections. For OntoNotes, gold segmentation is also available for the development and test sections.  On the other hand, no segmentation is available for the MSRA test sections, nor the Weibo / resume datasets. As a result, OntoNotes is leveraged for studying oracle situations where gold segmentation is given. We use the neural word segmentor of \citet{yang-zhang-dong:2017:Long} to automatically segment the development and test sets for word-based NER. In particular, for the OntoNotes and MSRA datasets, we train the segmentor using gold segmentation on their respective training sets. For Weibo and resume, we take the best model of \citet{yang-zhang-dong:2017:Long} off the shelf\footnote{\url{https://github.com/jiesutd/RichWordSegmentor}}, which is trained using CTB 6.0 \cite{xue2005penn}.

\textbf{Word Embeddings}. We pretrain word embeddings using word2vec \cite{mikolov2013distributed} over automatically segmented Chinese Giga-Word\footnote{\url{https://catalog.ldc.upenn.edu/LDC2011T13}}, obtaining 704.4k words in a final lexicon. In particular, the number of single-character, two-character and three-character words are 5.7k, 291.5k, 278.1k, respectively. The embedding lexicon is released alongside our code and models as a resource for further research. Word embeddings are fine-tuned during NER training. Character and character bigram embeddings are pretrained on Chinese Giga-Word using word2vec and fine-tuned at model training. 

\textbf{Hyper-parameter settings}. Table \ref{tab:hyperparameter} shows the values of hyper-parameters for our models, which as fixed according to previous work in the literature without grid-search adjustments for each individual dataset. In particular, the embedding sizes are set to 50 and the hidden size of LSTM models to 200. Dropout \cite{srivastava2014dropout} is applied to both word and character embeddings with a rate of 0.5. Stochastic gradient descent (SGD) is used for optimization, with an initial learning rate of 0.015 and a decay rate of 0.05.

\begin{table}[!tp]
\begin{center}
\resizebox{\columnwidth}{!}{%
\begin{tabular}{|l|l||l|l|}
\hline \textbf{Parameter} &  \textbf{Value} & \textbf{Parameter} & \textbf{Value}\\ \hline
char emb size & 50 & bigram emb size & 50 \\
lattice emb size &50 &LSTM hidden & 200\\
char dropout & 0.5 & lattice dropout & 0.5\\
LSTM layer &1  & regularization $\lambda$ &1e-8\\
learning rate \textit{lr} & 0.015 &\textit{lr} decay & 0.05\\
\hline
\end{tabular}
}
\end{center}
\caption{Hyper-parameter values.}
\label{tab:hyperparameter}
\end{table}

\subsection{Development Experiments}
We compare various model configurations on the OntoNotes development set, in order to select the best settings for word-based and character-based NER models, and to learn the influence of lattice word information on character-based models.

\textbf{Character-based NER}. As shown in Table \ref{tab:devsetting}, without using word segmentation, a character-based LSTM-CRF model gives a development F1-score of 62.47\%. Adding character-bigram and softword representations as described in Section \ref{sec:charmodel} increases the F1-score to 67.63\% and 65.71\%, respectively, demonstrating the usefulness of both sources of information. In addition, a combination of both gives a 69.64\% F1-score, which is the best among various character representations. We thus choose this model in the remaining experiments.

\textbf{Word-based NER}. Table \ref{tab:devsetting} shows a variety of different settings for word-based Chinese NER. With automatic segmentation, a word-based LSTM CRF baseline gives a 64.12\% F1-score, which is higher compared to the character-based baseline. This demonstrates that both word information and character information are useful for Chinese NER. The two methods of using character LSTM to enrich word representations in Section \ref{sec:word-based}, namely word+char LSTM and word+char LSTM$'$, lead to similar improvements.

A CNN representation of character sequences gives a slightly higher F1-score compared to LSTM character representations. On the other hand, further using character bigram information leads to increased F1-score over word+char LSTM, but decreased F1-score over word+char CNN. A possible reason is that CNN inherently captures character n-gram information. As a result, we use word+char+bichar LSTM for word-based NER in the remaining experiments, which gives the best development results, and is structurally consistent with the state-of-the-art English NER models in the literature.

\begin{figure}[!tp] 
  \centering 
  \includegraphics[width=2.6in]{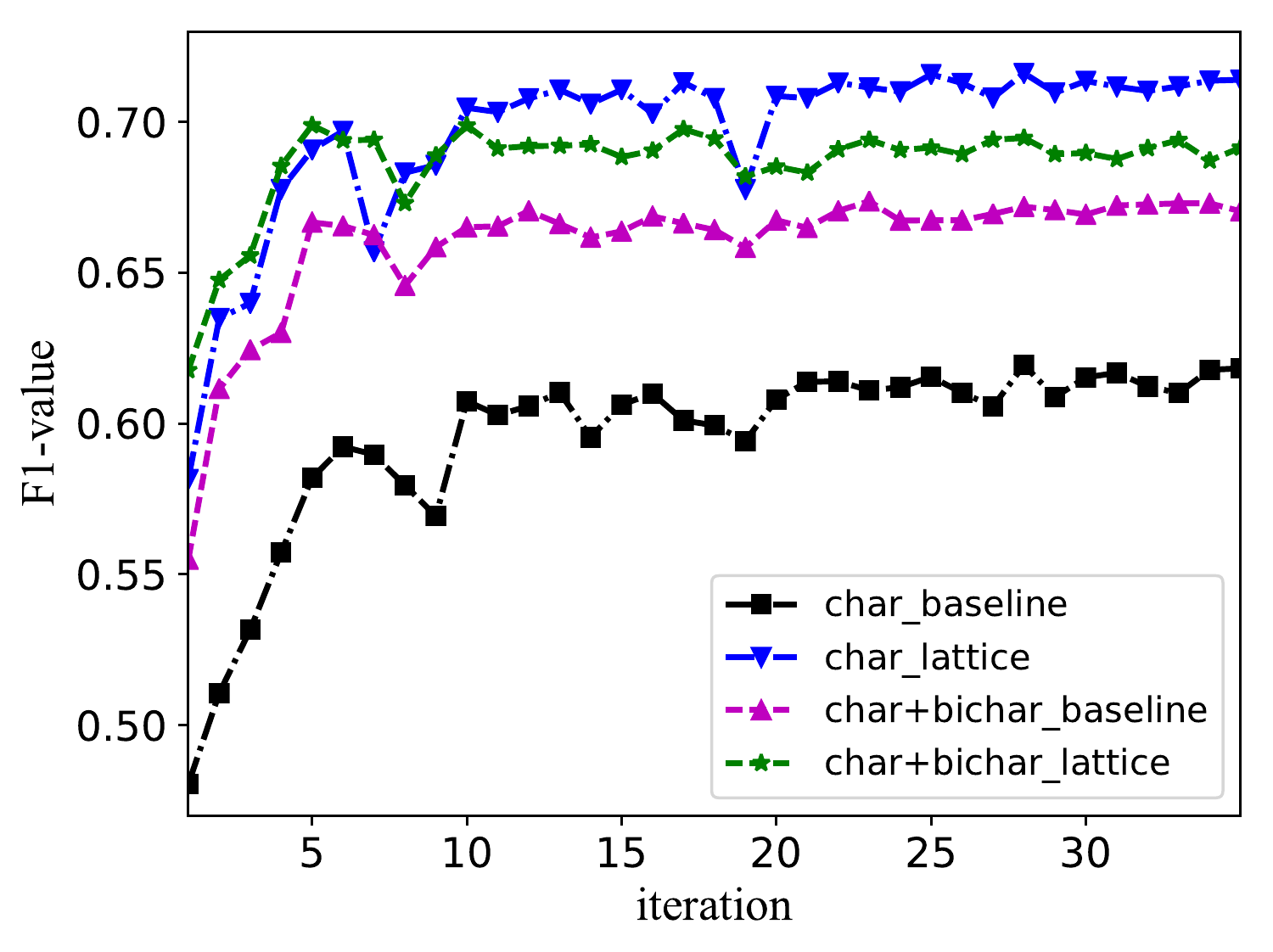}
  \vspace*{-5mm}
  \caption{F1 against training iteration number.}
  \label{fig:iteration}
\end{figure}

\begin{table}[!tp]
\begin{center}
\resizebox{\columnwidth}{!}{%
\begin{tabular}{|l|l|l|l|l|}
\hline 
\textbf{Input}&\textbf{Models}& \textbf{P} &\textbf{R} &\textbf{F1}  \\ 
\hline
\multirow{6}*{Auto seg}&Word baseline &73.20 &57.05 &64.12 \\
&\;\;\;+char LSTM&71.98 &65.41 &68.54 \\
&\;\;\;+char LSTM$'$ &71.08 &65.83 &68.35 \\
&\;\;\;+char+bichar LSTM &72.63 &67.60 &70.03 \\
&\;\;\;+char CNN&73.06 &66.29 &69.51 \\
&\;\;\;+char+bichar CNN&72.01 &65.50 &68.60 \\
\hline
\multirow{5}*{No seg}&Char baseline  &67.12 &58.42 &62.47 \\
&\;\;\;+softword &69.30 &62.47 &65.71\\
&\;\;\;+bichar &71.67 & 64.02 & 67.63\\
&\;\;\;+bichar+softword &72.64 &66.89 &69.64\\
\cline{2-5}
&Lattice 
 &\textbf{74.64} &\textbf{68.83} &\textbf{71.62} \\
\hline
\end{tabular}
}
\end{center}
\caption{Development results.}
\label{tab:devsetting}
\end{table}
\textbf{Lattice-based NER}. Figure \ref{fig:iteration} shows the F1-score of character-based and lattice-based models against the number of training iterations. We include models that use concatenated character and character bigram embeddings, where bigrams can play a role in disambiguating characters. As can be seen from the figure, lattice word information is useful for improving character-based NER, improving the best development result from 62.5\% to 71.6\%. On the other hand, the bigram-enhanced lattice model does not lead to further improvements compared with the original lattice model. This is likely because words are better sources of information for character disambiguation compared with bigrams, which are also ambiguous.

As shown in Table \ref{tab:devsetting}, the lattice LSTM-CRF model gives a development F1-score of 71.62\%, which is significantly\footnote{We use a $p$-value of less than 0.01 from pairwise t-test to indicate statistical significance.} higher compared with both the word-based and character-based methods, despite that it does not use character bigrams or word segmentation information. The fact that it significantly outperforms char+softword shows the advantage of lattice word information as compared with segmentor word information.

\begin{table}[!tp]
\begin{center}
\resizebox{\columnwidth}{!}{%
\begin{tabular}{|l|l|l|l|l|}
\hline 
\textbf{Input} &\textbf{Models}& \textbf{P} &\textbf{R} &\textbf{F1}  \\ 
\hline
\multirow{6}*{Gold seg}
&\citet{yang2017combining} &65.59 &71.84 &68.57\\
&\citet{yang2017combining}*\dag &72.98 &\textbf{80.15} &\textbf{76.40}\\
&\citet{che2013named}* &77.71 &72.51 &75.02\\
&\citet{wang2013effective}* &76.43 &72.32 &74.32\\
\cline{2-5}
&Word baseline &76.66 &63.60 &69.52\\
&\;\;\;+char+bichar LSTM &\textbf{78.62} &73.13 &75.77\\
\hline
\hline
\multirow{2}*{Auto seg}&Word baseline &72.84 &59.72&65.63\\
&\;\;\;+char+bichar LSTM &73.36 &70.12 &71.70\\
\hline
\multirow{3}*{No seg}&Char baseline  &68.79 &60.35 &64.30 \\
 &\;\;\;+bichar+softword &74.36 &69.43 &71.81 \\
 &Lattice &\textbf{76.35} &\textbf{71.56} &\textbf{73.88} \\
\hline
\end{tabular}
}
\end{center}
\caption{Main results on OntoNotes.}
\label{tab:ontotest}
\end{table}

\subsection{Final Results}
\textbf{OntoNotes}. The OntoNotes test results are shown in Table \ref{tab:ontotest}\footnote{In Table \ref{tab:ontotest}, \ref{tab:msratest} and \ref{tab:weibotest}, we use * to denote a model with external labeled data for semi-supervised learning. \dag ~ means that the model also uses discrete features.}. With gold-standard segmentation, our word-based methods give competitive results to the state-of-the-art on the dataset \cite{che2013named,wang2013effective}, which leverage bilingual data. This demonstrates that LSTM-CRF is a competitive choice for word-based Chinese NER, as it is for other languages. In addition, the results show that our word-based models can serve as highly competitive baselines. With automatic segmentation, the F1-score of word+char+bichar LSTM decreases from 75.77\% to 71.70\%, showing the influence of segmentation to NER. Consistent with observations on the development set, adding lattice word information leads to an 88.81\% $\rightarrow$ 93.18\% increasement of F1-score over the character baseline, as compared with 88.81\% $\rightarrow$ 91.87\% by adding bichar+softword. The lattice model gives significantly the best F1-score on automatic segmentation. 

\begin{table}[!tp]
\begin{center}
\scalebox{0.77}{
\begin{tabular}{|l|l|l|l|l|}
\hline 
\textbf{Models}& \textbf{P} &\textbf{R} &\textbf{F1}  \\ 
\hline
\citet{chen2006chinese} &91.22 &81.71 &86.20\\
\citet{zhang2006word}* &92.20 &90.18 &91.18 \\
\citet{zhou2013chinese} &91.86 &88.75 &90.28\\
\citet{lu2016multi} &-- & -- &87.94 \\
\citet{dong2016character} &91.28 &90.62 &90.95 \\
\hline
Word baseline &90.57 &83.06 &86.65 \\
\;\;\;+char+bichar LSTM &91.05 &89.53 &90.28\\
Char baseline  &90.74 &86.96 &88.81 \\
\;\;\;+bichar+softword& 92.97& 90.80& 91.87\\
Lattice &\textbf{93.57} &\textbf{92.79} &\textbf{93.18} \\
\hline
\end{tabular}
}
\end{center}
\caption{Main results on MSRA.}
\label{tab:msratest}
\end{table}

\begin{table}[!t]
\centering
\scalebox{0.77}{
\begin{tabular}{|l|l|l|l|}
\hline
{\textbf{Models}}&  \textbf{NE}& \textbf{NM}&\textbf{Overall}\\
\hline
\citet{peng2015named}&  51.96&  61.05& 56.05\\
\citet{peng2016improving}*& \textbf{55.28}& \textbf{62.97}& \textbf{58.99}\\
\citet{he2017f}&50.60& 59.32& 54.82\\
\citet{he2017unified}*& 54.50& 62.17& 58.23\\
\hline
Word baseline& 36.02& 59.38& 47.33\\
\;\;\;+char+bichar LSTM &43.40 &60.30 &52.33\\
Char baseline& 46.11&  55.29& 52.77\\
\;\;\;+bichar+softword& 50.55& 60.11& 56.75\\
Lattice &  \textbf{53.04} &\textbf{62.25}& \textbf{58.79}\\
\hline
\end{tabular}
}
\caption{Weibo NER results.}
\label{tab:weibotest}
\end{table}

\textbf{MSRA}. Results on the MSRA dataset are shown in Table \ref{tab:msratest}. For this benchmark, no gold-standard segmentation is available on the test set. Our chosen segmentor gives 95.93\% accuracy on 5-fold cross-validated training set. The best statistical models on the dataset leverage rich handcrafted features \cite{chen2006chinese,zhang2006word,zhou2013chinese} and character embedding features \cite{lu2016multi}. \citet{dong2016character} exploit neural LSTM-CRF with radical features. 

Compared with the existing methods, our word-based and character-based LSTM-CRF models give competitive accuracies. The lattice model significantly outperforms both the best character-based and word-based models ($p<0.01$), achieving the best result on this standard benchmark.

\textbf{Weibo/resume}. Results on the Weibo NER dataset are shown in Table \ref{tab:weibotest}, where NE, NM and Overall denote F1-scores for named entities, nominal entities (excluding named entities) and both, respectively. Gold-standard segmentation is not available for this dataset. Existing state-of-the-art systems include \citet{peng2016improving} and \citet{he2017unified}, who explore rich embedding features, cross-domain and semi-supervised data, some of which are orthogonal to our model\footnote{The results of \citet{peng2015named,peng2016improving} are taken from \citet{peng2017supplementary}.}.

Results on the resume NER test data are shown in Table \ref{tab:resumetest}. Consistent with observations on OntoNotes and MSRA, the lattice model significantly outperforms both the word-based mode and the character-based model for Weibo and resume ($p<0.01$), giving state-of-the-art results.

\begin{table}[!tp]
\begin{center}
\scalebox{0.77}{
\begin{tabular}{|l|l|l|l|l|}
\hline 
\textbf{Models}& \textbf{P} &\textbf{R} &\textbf{F1}  \\ 
\hline
Word baseline &93.72 &93.44 &93.58\\
\;\;\;+char+bichar LSTM &94.07 &94.42 &94.24\\
Char baseline  &93.66 &93.31 &93.48 \\
\;\;\;+bichar+softword& 94.53& \textbf{94.29}& 94.41\\
Lattice &\textbf{94.81} &94.11 &\textbf{94.46} \\
\hline
\end{tabular}
}
\end{center}
\caption{Main results on resume NER.}
\label{tab:resumetest}
\end{table}

\begin{figure}[!tp] 
  \centering 
  \includegraphics[width=2.6in]{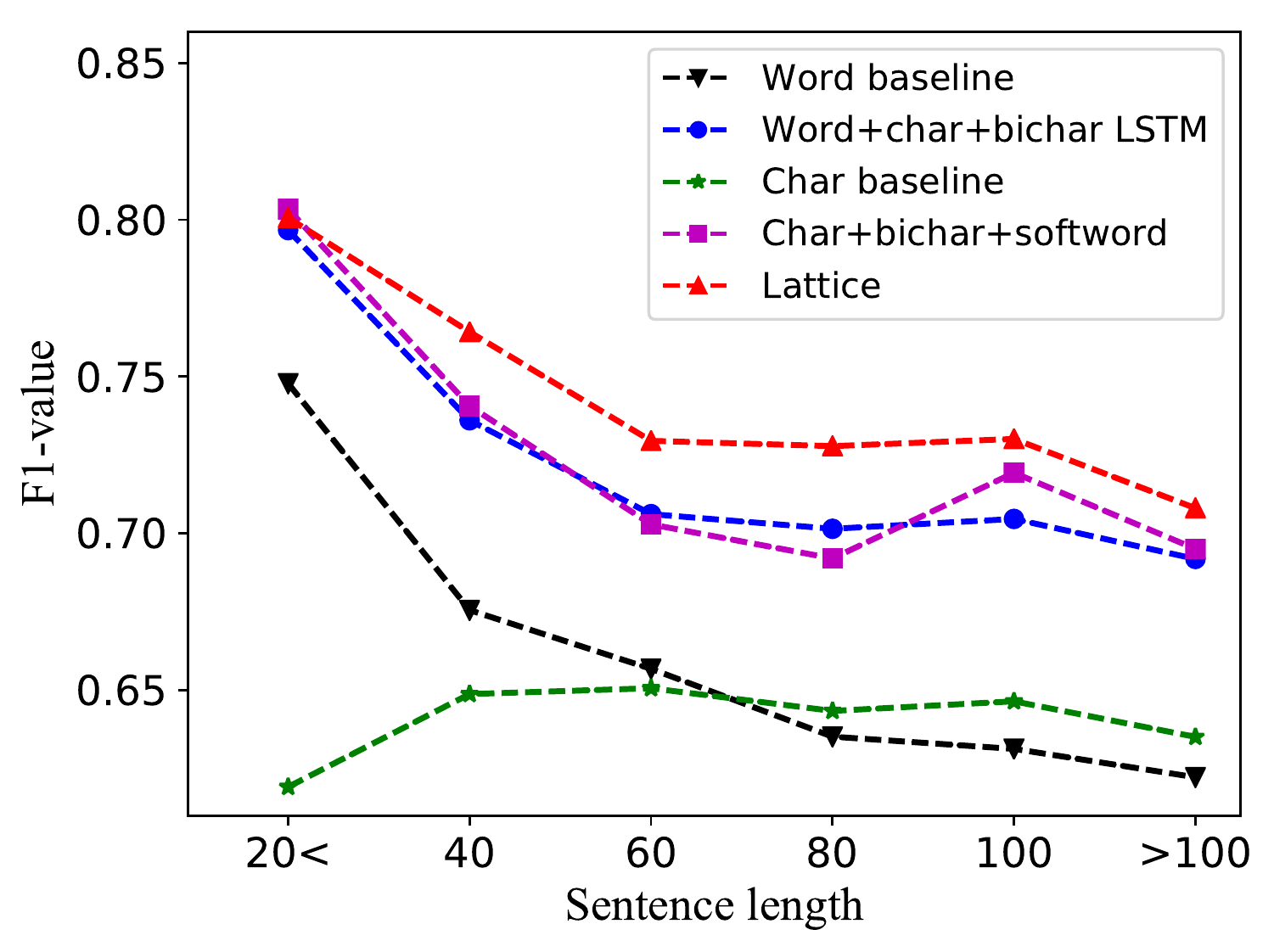}
  \vspace*{-5mm}
  \caption{F1 against sentence length.}
  \label{fig:sentLength}
\end{figure}

\subsection{Discussion}
\noindent \textbf{F1 against sentence length}. Figure \ref{fig:sentLength} shows the F1-scores of the baseline models and lattice LSTM-CRF on the OntoNotes dataset. The character-based baseline gives relatively stable F1-scores over different sentence lengths, although the performances are relatively low. The word-based baseline gives substantially higher F1-scores over short sentences, but lower F1-scores over long sentences, which can be because of lower segmentation accuracies over longer sentences. Both word+char+bichar and char+bichar+softword give better performances compared to their respective baselines, showing that word and character representations are complementary for NER. The accuracy of lattice also decreases as the sentence length increases, which can result from exponentially increasing number of word combinations in lattice. Compared with word+char+bichar and char+bichar+softword, the lattice model shows more robustness to increased sentence lengths, demonstrating the more effective use of word information. 

\begin{table}[!t] 
  \resizebox{\columnwidth}{!}{%
  \begin{tabular}{|l|l|}
  \hline
  \multirow{2}*{\textbf{Sentence} (truncated)}& 卸下东莞台协会长职务后\\
  & After stepping down as president of Taiwan Association in Dongguan.\\
  \hline
  \multirow{2}*{\textbf{Correct Segmentation}} &卸下\;\;东莞\;\;台\;\;协\;\;会长\;\;职务\;\;后\\
  &step down, Dongguan, Taiwan, association, president, role, after\\
  \hline
  \multirow{2}*{\textbf{Auto Segmentation}} &卸下\;\;东莞\;\;台\;\;协会长\;\;职务\;\;后\\
  &step down, Dongguan, Taiwan, association president, role, after\\
  \hline
  \multirow{3}*{\textbf{Lattice words}} &卸下\;\;下东\;\;东莞\;\;台协会\;\;协会\;\;会长\;\;长职\;\;职务\\
  & step down, incorrect word, Dongguan, Taiwan association, \\
  &association, president, permanent job, role\\
  \hline
  \hline
  \multirow{2}*{\textbf{Word+char+bichar LSTM}}&卸下\colorbox{red!30}{东莞}$_{GPE}$\colorbox{red!30}{台}$_{GPE}$协会长职务后\\
  &\ldots \colorbox{red!30}{Dongguan}$_{GPE}$ \colorbox{red!30}{Taiwan}$_{GPE}$ \ldots\\
  \hline
  \multirow{2}*{\textbf{Char+bichar+softword}}&卸下\colorbox{red!30}{东莞台协会}$_{ORG}$长职务后\\
  &\ldots \colorbox{red!30}{Taiwan Association in Dongguan}$_{ORG}$ \ldots (ungrammatical) \\
  \hline
  \multirow{2}*{\textbf{Lattice}}&卸下\colorbox{green!30}{东莞台协}$_{ORG}$会长职务后\\
  &\ldots \colorbox{green!30}{Taiwan Association in Dongguan}$_{ORG}$ \ldots\\
  \hline
  \end{tabular}
  }
  \caption{Example. Red and green represent incorrect and correct entities, respectively.}
  \label{tab:resultexample}
\end{table}

\noindent \textbf{Case Study}. Table \ref{tab:resultexample} shows a case study comparing char+bichar+softword, word+char+bichar and the lattice model. In the example, there is much ambiguity around the named entity ``东莞台协\;(Taiwan Association in Dongguan)''. Word+char+bichar yields the entities ``东莞\;(Dongguan)'' and ``台\;(Taiwan)'' given that ``东莞台协\;(Taiwan Association in Dongguan)'' is not in the segmentor output. Char+bichar+softword recognizes ``东莞台协会\;(Taiwan Association in Dongguan)'', which is valid on its own, but leaves the phrase ``长职务后'' ungrammatical. In contrast, the lattice model detects the organization name correctly, thanks to the lattice words ``东莞\;(Dongguan)'', ``会长\;(President)'' and ``职务\;(role)''. There are also irrelevant words such as ``台协会\;(Taiwan Association)'' and ``下东\;(noisy word)'' in the lexicon, which did not affect NER results.

Note that both word+char+bichar and lattice use the same source of word information, namely the same pretrained word embedding lexicon. However, word+char+bichar first uses the lexicon in the segmentor, which imposes hard constrains (i.e. fixed words) to its subsequence use in NER. In contrast, lattice LSTM has the freedom of considering all lexicon words.

\noindent \textbf{Entities in lexicon}. Table \ref{tab:lexiconmatch} shows the total number of entities and their respective match ratios in the lexicon. The error reductions (ER) of the final lattice model over the best character-based method (i.e. ``+bichar+softword'') are also shown. It can be seen that error reductions have a correlation between matched entities in the lexicon.  In this respect, our automatic lexicon also played to some extent the role of a gazetteer \cite{ratinov2009design,chiu2015named}, but not fully since there is no explicit knowledge in the lexicon which tokens are entities. The ultimate disambiguation power still lies in the lattice encoder and supervised learning. 

The quality of the lexicon may affect the accuracy of our NER model since noise words can potentially confuse NER. On the other hand, our lattice model can potentially learn to select more correct words during NER training. We leave the investigation of such influence to future work.

\begin{table}[!tp]
\begin{center}
\resizebox{\columnwidth}{!}{%
\begin{tabular}{|l|l|l|l|l|l|}
\hline 
\textbf{Dataset} &\textbf{Split}&\textbf{\#Entity}  & \textbf{\#Match} & \textbf{Ratio} (\%) & \textbf{ER} (\%)\\ 
\hline
\multirow{2}*{OntoNotes} 
&Train &13.4k  &9.5k  &71.04 &--\\
&Test &7.7k &6.0k &78.72 &7.34\\
\hline
\multirow{2}*{MSRA} 
&Train &74.7k   &54.3k  &72.62 &--\\
&Test &6.2k   &4.6k &73.76&16.11\\
\hline
\multirow{2}*{Weibo (all)} 
&Train &1.9k   &1.1k  &58.83 &--\\
&Test & 414   &259 &62.56&4.72\\
\hline
\multirow{2}*{resume} 
&Train &13.4k   &3.8k &28.55 &--\\
&Test & 1.6k   &483 &29.63 &0.89\\
\hline
\end{tabular}
}
\end{center}
\caption{Entities in lexicon.}
\label{tab:lexiconmatch}
\end{table}

\section{Conclusion}
We empirically investigated a lattice LSTM-CRF representations for Chinese NER, finding that it gives consistently superior performance compared to word-based and character-based LSTM-CRF across different domains. The lattice method is fully independent of word segmentation, yet more effective in using word information thanks to the freedom of choosing lexicon words in a context for NER disambiguation.

\section*{Acknowledgments}
We thank the anonymous reviewers for their insightful comments.

\bibliography{acl2018}
\bibliographystyle{acl_natbib}

\end{CJK*}
\end{document}